\documentclass{article}

\usepackage{microtype}
\usepackage{graphicx}
\usepackage{subfigure}
\usepackage{booktabs}
\usepackage{amsfonts}
\usepackage{bm}
\usepackage{setspace}

\renewcommand{\mathbf}{\boldsymbol}

\usepackage{amsmath,amssymb,amsthm}
\usepackage{multirow}
\newtheorem{theorem}{Theorem}
\newtheorem{definition}[theorem]{Definition}
\newtheorem{lemma}[theorem]{Lemma}
\newtheorem{corollary}[theorem]{Corollary}

\usepackage{hyperref}

\usepackage[accepted]{icml2020}

\icmltitlerunning{Analytic Marching: An Analytic Meshing Solution from Deep Implicit Surface Networks}

\begin{document}

\twocolumn[
\icmltitle{Analytic Marching: An Analytic Meshing Solution from \\ Deep Implicit Surface Networks}



\icmlsetsymbol{equal}{*}

\begin{icmlauthorlist}
    \icmlauthor{Jiabao Lei}{scut}
    \icmlauthor{Kui Jia}{scut}
\end{icmlauthorlist}

\icmlaffiliation{scut}{School of Electronic and Information Engineering,
                        South China University of Technology,
                        Guangzhou, Guangdong, China}

\icmlcorrespondingauthor{Kui Jia}{kuijia@scut.edu.cn}

\icmlkeywords{Mesh reconstruction, implicit surface function, deep learning}

\vskip 0.3in
]



\printAffiliationsAndNotice  


\begin{abstract}
This paper studies a problem of learning surface mesh via implicit functions in an emerging field of deep learning surface reconstruction, where implicit functions are popularly implemented as multi-layer perceptrons (MLPs) with rectified linear units (ReLU). To achieve meshing from learned implicit functions, existing methods adopt the de-facto standard algorithm of marching cubes; while promising, they suffer from loss of precision learned in the MLPs, due to the discretization nature of marching cubes. Motivated by the knowledge that a ReLU based MLP partitions its input space into a number of linear regions, we identify from these regions \emph{analytic cells} and \emph{analytic faces} that are associated with zero-level isosurface of the implicit function, and characterize the theoretical conditions under which the identified analytic faces are guaranteed to connect and form a \emph{closed, piecewise planar surface}. Based on our theorem, we propose a naturally parallelizable algorithm of \emph{analytic marching}, which marches among analytic cells to \emph{exactly} recover the mesh captured by a learned MLP. Experiments on deep learning mesh reconstruction verify the advantages of our algorithm over existing ones.

\end{abstract}

\section{Introduction}

This paper studies a geometric notion of object surface whose nature is a 2-dimensional manifold embedded in the 3D space. In literature, there exist many different ways to represent an object surface, either explicitly or implicitly \cite{MarioBotschPolygonalMesh}. For example, one of the most popular representations is \emph{polygonal mesh} that approximates a smooth surface as piecewise linear functions. Mesh representation of object surface plays fundamental roles in many applications of computer graphics and geometry processing, e.g., computer-aided design, movie production, and virtual/augmented reality.

As a parametric representation of object surface, the most typical \emph{triangle mesh} is explicitly defined as a collection of connected faces, each of which has three vertices that uniquely determine plane parameters of the face in the 3D space. However, parametric surface representations are usually difficult to obtain, especially for topologically complex surface; queries of points inside or outside the surface are expensive as well. Instead, one usually resorts to implicit functions (e.g., signed distance function or SDF \cite{TSDFCurlessL96,DeepSDF}), which subsume the surface as zero-level isosurface in the function field; other implicit representations include discrete volumes and those based on algebraic \cite{blobby_molecules,metaballs,soft_object} and radial basis functions \cite{RBF1,RBF2,RBF3}. To obtain a surface mesh, the continuous field function is often discretized around the object as a regular grid of voxels, followed by the de-facto standard algorithm of marching cubes \cite{MC}. Efficiency and result regularity of marching cubes can be improved on a hierarchically sampled structure of octree via algorithms such as dual contouring \cite{DC}.

The most popular implicit function of SDF is traditionally implemented discretely as a regular grid of voxels. More recently, methods of deep learning surface reconstruction \cite{DeepSDF,DISN} propose to use deep models of Multi-Layer Perceptron (MLP) to learn continuous SDFs; given a learned SDF, they typically take a final step of marching cubes to obtain the mesh reconstruction results. While promising, the final step of marching cubes recovers a mesh that is only an approximation of the surface captured by the learned SDF; more specifically, it suffers from a trade-off of efficiency and precision, due to the discretization nature of the marching cubes algorithm.

Motivated by the established knowledge that an MLP based on rectified linear units (ReLU) \cite{ReLU} partitions its input space into a number of linear regions \cite{Montufar2014}, we identify from these regions \emph{analytic cells} and \emph{analytic faces} that are associated with zero-level isosurface of the MLP based SDF. Assuming that such a SDF learns its zero-level isosurface as a \emph{closed, piecewise planar surface}, we characterize theoretical conditions under which analytic faces of the SDF \emph{connect and exactly form} the surface mesh. Based on our theorem, we propose an algorithm of \emph{analytic marching}, which marches among analytic cells to recover the \emph{exact mesh} of the closed, piecewise planar surface captured by a learned MLP. Our algorithm can be naturally implemented in parallel. We present careful ablation studies in the context of deep learning mesh reconstruction. Experiments on benchmark datasets of 3D object repositories show the advantages of our algorithm over existing ones, particularly in terms of a better trade-off of precision and efficiency.

\section{Related works}

The problem studied in this work is closely related to the following three lines of research.

\noindent\textbf{Implicit surface representation} To represent an object surface implicitly, some of previous methods take a strategy of divide and conquer that represents the surface using atom functions. For example, blobby molecule \cite{blobby_molecules} is proposed to approximate each atom by a gaussian potential, and a piecewise quadratic meta-ball \cite{metaballs} is used to approximate the gaussian, which is improved via a soft object model in \cite{soft_object} by using a sixth degree polynomial. Radial basis function (RBF) is an alternative to the above algebraic functions. RBF-based approaches \cite{RBF1,RBF2,RBF3} place the function centers near the surface and are able to reconstruct a surface from a discrete point cloud, where common choices of basis function include thin-plate spline, gaussian, multiquadric, and biharmonic/triharmonic splines.

\noindent\textbf{Methods of mesh conversion} The conversion from an implicit representation to an explicit surface mesh is called isosurface extraction. Probably the simplest approach is to directly convert a volume via greedy meshing (GM). The de-facto standard algorithm of marching cubes (MC) \cite{MC} builds from the implicit function a discrete volume around the object, and computes mesh vertices on the edges of the volume; due to its discretization nature, mesh results of the algorithm are often short of sharp surface details. Algorithms similar to MC include marching tetrahedra (MT) \cite{MT} and dual contouring (DC) \cite{DC}. In particular, MT divides a voxel into six tetrahedrons and calculates the vertices on edges of each tetrahedron; DC utilizes gradients to estimate positions of vertices in a cell and extracts meshes from adaptive octrees. All these methods suffer from a trade-off of precision and efficiency due to the necessity to sample discrete points from the 3D space.

\noindent\textbf{Local linearity of MLPs} Among the research studying representational complexities of deep networks, {Mont{\'u}far} et al. \yrcite{Montufar2014} and Pascanu et al. \yrcite{pascanu2013number} investigate how a ReLU or maxout based MLP partitions its input space into a number of linear regions, and bound this number via quantities relevant to network depth and width. The region-wise linear mapping is explicitly established in \cite{OrthDNNs} in order to analyze generalization properties of deep networks. A closed-form solution termed OpenBox is proposed in \cite{OpenBox} that computes consistent and exact interpretations for piecewise linear deep networks. The present work leverages the locally linear properties of MLP networks and studies how zero-level isosurface can be identified from MLP based SDFs.

\section{Analytic meshing via deep implicit surface networks}

We start this section with the introduction of Multi-Layer Perceptrons (MLPs) based on rectified linear units (ReLU) \cite{ReLU}, and discuss how such an MLP as a nonlinear function partitions its input space into linear regions via a compositional structure.


\subsection{Local linearity of multi-layer perceptions}\label{SecMLPAnalysis}

An MLP of $L$ hidden layers takes an input $\mathbf{x} \in \mathbb{R}^{n_0}$ from the space $\mathcal{X}$, and layer-wisely computes $\mathbf{x}_l = \mathbf{g} (\mathbf{W}_l \mathbf{x}_{l-1})$, where $l \in \{1, \dots, L\}$ indexes the layer, $\mathbf{x}_l \in \mathbb{R}^{n_l}$, $\mathbf{x}_0 = \mathbf{x}$, $\mathbf{W}_l \in \mathbb{R}^{n_l\times n_{l-1}}$, and $\mathbf{g}$ is the point-wise, ReLU based activation function. We also denote the intermediate feature space $g(\mathbf{W}_l\mathbf{x}_{l-1})$ as $\mathcal{X}_l$ and $\mathcal{X}_0 = \mathcal{X}$. We compactly write the MLP as a mapping $\mathbf{T}\mathbf{x} = \mathbf{g}(\mathbf{W}_{L}\ldots \mathbf{g}(\mathbf{W}_1\mathbf{x}))$. Any $k^{th}$ neuron, $k \in \{1,\ldots, n_l\}$, of an $l^{th}$ layer of the MLP $\mathbf{T}$ specifies a functional defined as
\begin{displaymath}
a_{lk}(\mathbf{x}) = \pi_k\mathbf{g}(\mathbf{W}_l\mathbf{g}(\mathbf{W}_{l-1}\ldots \mathbf{g}(\mathbf{W}_1\mathbf{x}))) ,
\end{displaymath}
where $\pi_k$ denotes an operator that projects onto the $k^{th}$ coordinate. All the neurons at layer $l$ define a functional as
\begin{displaymath}
\mathbf{a}_{l}(\mathbf{x}) =  \mathbf{g}(\mathbf{W}_l\mathbf{g}(\mathbf{W}_{l-1}\ldots \mathbf{g}(\mathbf{W}_1\mathbf{x}))) .
\end{displaymath}
We define the support of $\mathbf{T}$ as
\begin{equation}\label{EqnMLPSupport}
\textrm{supp}(\mathbf{T}) = \{ \mathbf{x} \in \mathcal{X} | \mathbf{T}\mathbf{x} \not= \mathbf{0}\} ,
\end{equation}
which are instances of practical interest in the input space. Support $\textrm{supp}(a_{lk})$ of any neuron $a_{lk}$ is similarly defined.

For an intermediate feature space $\mathcal{X}_{l-1} \in \mathbb{R}^{n_{l-1}}$, each hidden neuron of layer $l$ specifies a hyperplane $H$ that partitions $\mathcal{X}_{l-1}$ into two halves, and the collection of hyperplanes $\{H_i\}_{i=1}^{n_l}$ specified by all the $n_l$ neurons of layer $l$ form a \emph{hyperplane arrangement} \cite{orlik1992arrangements}. These hyperplanes partition the space $\mathcal{X}_{l-1}$ into multiple linear regions whose formal definition is as follows.
\begin{definition}[Region/Cell]
Let $\mathcal{A}$ be an arrangement of hyperplanes in $\mathbb{R}^m$. A region of the arrangement is a connected component of the complement $\mathbb{R}^m - \bigcup\limits_{H \in \mathcal{A}}H$. A region is a cell when it is bounded. 
\end{definition}
Classical result from \cite{Zaslavsky1975,pascanu2013number} tells that the arrangement of $n_l$ hyperplanes  gives at most $\sum_{j=0}^{n_{l-1}}{{n_l}\choose{j}}$ regions in $\mathbb{R}^{n_{l-1}}$. Given fixed $\{\mathbf{W}_l\}_{l=1}^{L}$, the MLP $\mathbf{T}$ partitions the input space $\mathcal{X} \in \mathbb{R}^{n_0}$ by its layers' recursive partitioning of intermediate feature spaces, which can be intuitively understood as a successive process of space folding \cite{Montufar2014}.

Let $\mathcal{R}(\mathbf{T})$, shortened as $\mathcal{R}$, denote the set of all linear regions/cells in $\mathbb{R}^{n_0}$ that are possibly achieved by $\mathbf{T}$. To have a concept on the maximal size of $\mathcal{R}$, we introduce the following functionals about activation states of neuron, layer, and the whole MLP.
\begin{definition}[State of Neuron/MLP]\label{DefinitionNeuronMLPState}
For a $k^{th}$ neuron of an $l^{th}$ layer of an MLP $\mathbf{T}$, with $k \in \{1, \dots, n_l\}$ and $l \in \{1, \dots, L\}$, its state functional of neuron activation is defined as
\begin{equation}\label{EqnNeuronStateFunctional}
s_{lk}(\mathbf{x}) =
\begin{cases}
1 & \text{if $a_{lk}(\mathbf{x}) > 0$}\\
0 & \text{if $a_{lk}(\mathbf{x}) \leq 0$} ,
\end{cases}
\end{equation}
which gives the state functional of layer $l$ as
\begin{equation}\label{EqnLayerStateFunctional}
\mathbf{s}_l(\mathbf{x}) = [s_{l1}(\mathbf{x}), \dots, s_{ln_l}(\mathbf{x})]^{\top} ,
\end{equation}
and the state functional of MLP $\mathbf{T}$ as
\begin{equation}\label{EqnMLPStateFunctional}
\mathbf{s}(\mathbf{x}) = [s_1(\mathbf{x})^{\top}, \dots, s_L(\mathbf{x})^{\top}]^{\top} .
\end{equation}
\end{definition}
Let the total number of hidden neurons in $\mathbf{T}$ be $N = \sum_{l=1}^L n_l$. Denote $\mathbb{J} = \{1, 0\}$, and we have the state functional $\mathbf{s} \in \mathbb{J}^N$. Considering that a region in $\mathbb{R}^{n_0}$ corresponds to a realization of $\mathbf{s} \in \mathbb{J}^N$, it is clear that the maximal size of $\mathcal{R}$ is upper bounded by $2^N$. This gives us the following labeling scheme: for any region $r \in \mathcal{R}$, it corresponds to a unique element in $\mathbb{J}^N$; since $\mathbf{s}(\mathbf{x})$ is fixed for all $\mathbf{x} \in \mathcal{X}$ that fall in a same region $r$,  we use $\mathbf{s}(r) \in \mathbb{J}^N$ to label this region. Results from \cite{Montufar2014} tell that when layer widths of $\mathbf{T}$ satisfy $n_l \geq n_0$ for any $l \in \{1, \dots, L\}$, the maximal size of $\mathcal{R}(\mathbf{T})$ is lower bounded by $\left( \prod_{l=1}^{L-1} \lfloor n_l/n_0 \rfloor^{n_0} \right) \sum_{j=0}^{n_0}{{n_L}\choose{j}}$, where $\lfloor\cdot\rfloor$ ignores the remainder. Assuming $n_1 = \cdots = n_L = n$, the lower bound has an order of $\mathcal{O}\left( (n/n_0)^{(L-1)n_0} n^{n_0}  \right)$, which grows exponentially with the network depth and polynomially with the network width.
We have the following lemma adapted from \cite{OrthDNNs} to characterize the region-wise linear mappings.
\begin{lemma}[Linear Mapping of Region/Cell, an adaptation of Lemma 3.2 in \cite{OrthDNNs}]
\label{LemmaRegionMapping}
Given a ReLU based MLP $\mathbf{T}$ of $L$ hidden layers, for any region/cell $r\in \mathcal{R}(\mathbf{T})$, its associated linear mapping $\mathbf{T}^r$ is defined as
\begin{align}
  \mathbf{T}^r &= \prod\limits_{l=1}^{L}\mathbf{W}^r_l \label{EqnMLPRegionMapping} \\
  \mathbf{W}^r_{l} &= \textrm{diag}(\mathbf{s}_l(r))\mathbf{W}_{l} \label{EqnMLPRegionMappingLayerComp},
\end{align}
where $\textrm{diag}(\cdot)$ diagonalizes the state vector $\mathbf{s}_l(r)$.
\end{lemma}
Intuitively, the state vector $\mathbf{s}_l$ in (\ref{EqnMLPRegionMappingLayerComp}) selects a submatrix from $\mathbf{W}_{l}$ by setting those inactive rows as zero.

\subsection{Analytic cells and faces associated with the zero level isosurface of an implicit field function}

Let $F: \mathbb{R}^3 \rightarrow \mathbb{R}$ denote a scalar-valued, implicit field of signed distance function (SDF). Given $F$, an object surface $\mathcal{Z}$ is formally defined as its zero-level isosurface, i.e., $\mathcal{Z} = \{ \mathbf{x} \in \mathbb{R}^3 | F(\mathbf{x}) = 0 \}$. We also have the distance $F(\mathbf{x}) > 0$ for points inside $\mathcal{Z}$ and $F(\mathbf{x}) < 0$ for those outside. While $F$ can be realized using radial basis functions \cite{RBF1,RBF2,RBF3} or be approximated as a regular grid of voxels (i.e., a volume), in this work, we are particularly interested in implementing $F$ using ReLU based MLPs, which become an increasingly popular choice in recent works of deep learning surface reconstruction \cite{DeepSDF,DISN}.

\begin{figure}[!htbp]
    \vskip 0.1in
    \begin{center}
    \includegraphics[scale=0.3]{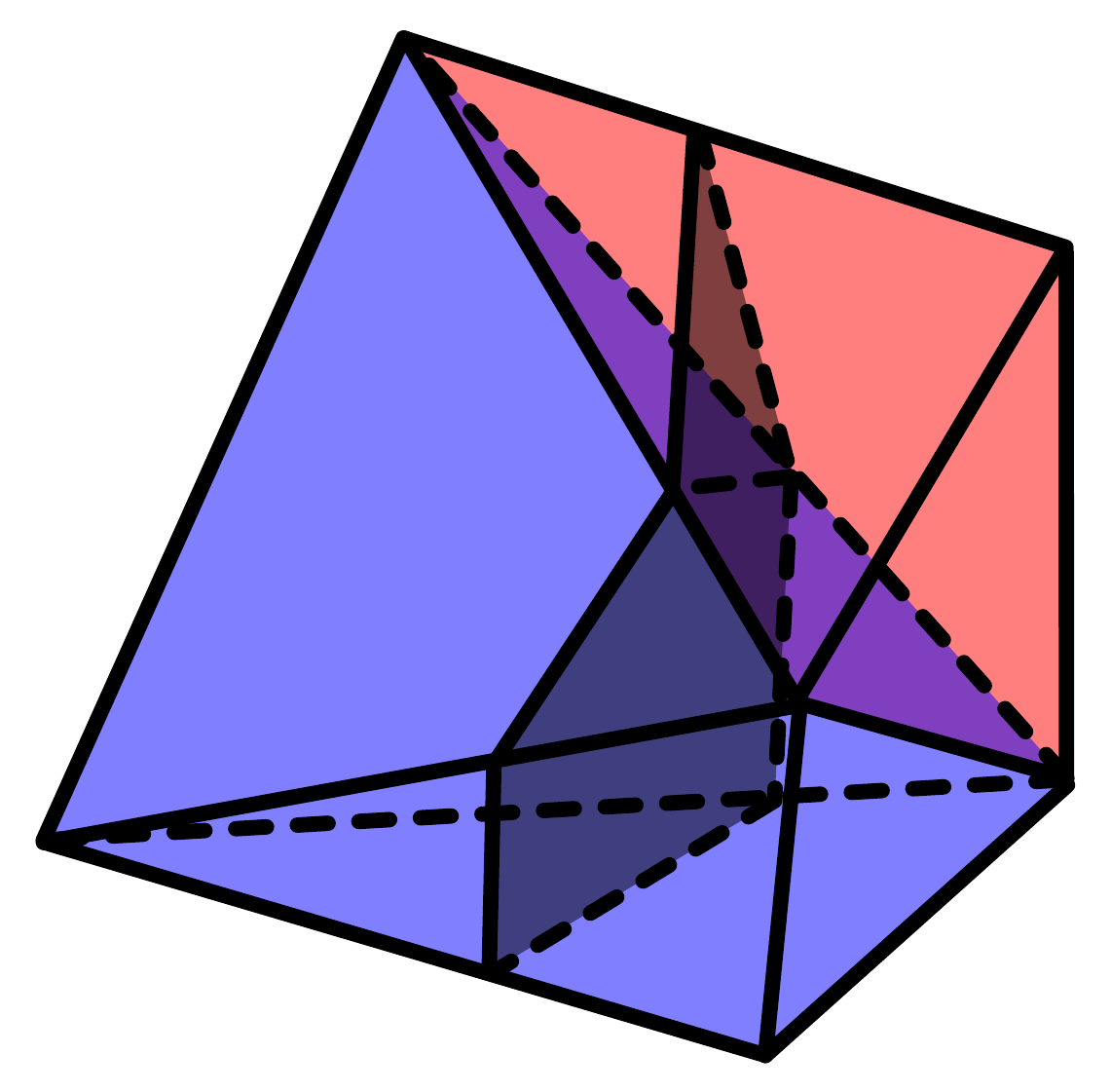} 
    \caption{An illustration that two analytic cells (respectively colored as blue and red) connect via a shared boundary plane, on which their associated analytic faces intersect to form a mesh edge. }
    \label{FigAnalyticCellFaceIllus}
    \end{center}
    \vskip -0.1in
\end{figure}

To implement $F$ using an MLP $\mathbf{T}$, we stack on top of $\mathbf{T}$ a regression function $f: \mathbb{R}^{n_L} \rightarrow \mathbb{R}$, giving rise to a functional of SDF as
\begin{displaymath}
F(\mathbf{x}) = f\circ\mathbf{T}(\mathbf{x}) = \mathbf{w}_f^{\top}\mathbf{g}(\mathbf{W}_{L}\ldots \mathbf{g}(\mathbf{W}_1\mathbf{x})) ,
\end{displaymath}
where $\mathbf{w}_f \in \mathbb{R}^{n_L}$ is weight vector of the regressor. Since $F$ represents a field function in the 3D Euclidean space, we have $n_0 = 3$. Analysis in Section \ref{SecMLPAnalysis} tells that the MLP $\mathbf{T}$ partitions the input space $\mathbb{R}^3$ into a set $\mathcal{R}$ of linear regions; any region $r \in \mathcal{R}$ that satisfies $\mathbf{x} \in \text{supp}(\mathbf{T}) \ \forall \ \mathbf{x} \in  r$ can be uniquely indexed by its state vector $\mathbf{s}(r)$ defined by (\ref{EqnMLPStateFunctional}). For such a region $r$, we have the following corollary from Lemma \ref{LemmaRegionMapping} that characterizes the associated linear mappings defined at neurons of $\mathbf{T}$ and the final regressor.
\begin{corollary}\label{LemmaRegionAssociatedNeuronMapping}
Given a SDF $F = f\circ\mathbf{T}$ built on a ReLU based MLP of $L$ hidden layers, for any $r\in \mathcal{R}(\mathbf{T})$, the associated neuron-wise linear mappings and that of the final regressor are defined as
\begin{align}
\mathbf{a}^r_{lk} &= \pi_{k}\prod\limits_{i=1}^{l}\mathbf{W}^r_{i} \label{EqnRegionAssociatedNeuronMapping} \\
\mathbf{a}_F^r &= \mathbf{w}_f^{\top}\mathbf{T}^r = \mathbf{w}_f^{\top}\prod\limits_{i=1}^{L}\mathbf{W}^r_{i} \label{EqnSDFPlaneFunctional} ,
\end{align}
where $l \in \{1, \dots, L\}$, $k \in \{1, \dots, n_l\}$, and $\mathbf{T}^r$ and $\mathbf{W}^r_{i}$ are defined as in Lemma \ref{LemmaRegionMapping}.
\end{corollary}
Corollary \ref{LemmaRegionAssociatedNeuronMapping} tells that the SDF $F$ in fact induces a set of region-associated planes in $\mathbb{R}^3$. The plane $\{\mathbf{x} \in \mathbb{R}^3 | \mathbf{a}_F^r\mathbf{x} = 0\}$ and the associated region $r$ have the following relations, assuming that they are in general positions. In the following, we use the normal $\mathbf{a}_F^r$ to represent the plane for simplicity.
\begin{itemize}
\item \emph{Intersection} $\mathbf{a}_F^r$ splits the region $r$ into two halves, denoted as $r^{+}$ and $r^{-}$, such that $\forall \mathbf{x} \in r^{+}$, we have $\mathbf{a}_F^r\mathbf{x} > 0$ and $\forall \mathbf{x} \in r^{-}$, we have $\mathbf{a}_F^r\mathbf{x} \leq 0$.
\item \emph{Non-intersection} We either have $\mathbf{a}_F^r\mathbf{x} > 0$ or $\mathbf{a}_F^r\mathbf{x} < 0$ for all $\mathbf{x} \in r$.
\end{itemize}
Let $\{\tilde{r} \in \widetilde{\mathcal{R}}\}$ denote the subset of regions in $\mathcal{R}$ that have the above relation of intersection, an illustration of which is given in Figure \ref{FigAnalyticCellFaceIllus}. It is clear that the zero-level isosurface $\mathcal{Z} = \{ \mathbf{x} \in \mathbb{R}^3 | F(\mathbf{x}) = 0 \}$ defined on the support (\ref{EqnMLPSupport}) of $\mathbf{T}$ can be only in $\widetilde{\mathcal{R}}$. To have an analytic understanding on any $\tilde{r} \in \widetilde{\mathcal{R}}$, we note from Corollary \ref{LemmaRegionAssociatedNeuronMapping} that the boundary planes of $\tilde{r}$ must be among the set
\begin{equation}\label{EqnBoundaryPlanes}
\{ H_{lk}^{\tilde{r}}\} \ \textrm{s.t.} \ H_{lk}^{\tilde{r}} = \{ \mathbf{x} \in \mathbb{R}^3 | \mathbf{a}^{\tilde{r}}_{lk}\mathbf{x} = 0 \} ,
\end{equation}
where $l = 1, \dots, L$ and $k = 1, \dots, n_l$;
for any $\mathbf{x} \in \tilde{r}$, it must satisfy $\textrm{sign}(\mathbf{a}^{\tilde{r}}_{lk}\mathbf{x}) = (2 s_{lk}(\tilde{r}) - 1)$, which gives the following system of inequalities
\begin{equation}\label{EqnAnalyticCellSystem}
(\mathbf{I} - 2\textrm{diag}(\mathbf{s}(\tilde{r})) \mathbf{A}^{\tilde{r}}\mathbf{x}  =
\begin{bmatrix}
(1 - 2s_{11}(\tilde{r}))\mathbf{a}^{\tilde{r}}_{11} \\
\vdots \\
(1 - 2s_{lk}(\tilde{r}))\mathbf{a}^{\tilde{r}}_{lk} \\
\vdots \\
(1 - 2s_{Ln_L}(\tilde{r}))\mathbf{a}^{\tilde{r}}_{Ln_L}
\end{bmatrix}
\mathbf{x} \preceq 0 ,
\end{equation}
where $\mathbf{I}$ is an identity matrix of compatible size, $\mathbf{A}^{\tilde{r}} \in \mathbb{R}^{Ln_L\times 3}$ collects the coefficients of the inequalities, and the state functionals $s_{lk}$ and $\mathbf{s}$ are defined by (\ref{EqnNeuronStateFunctional}) and (\ref{EqnMLPStateFunctional}).
We note that for some cases of region $\tilde{r}$, there could exist redundance in the inequalities of (\ref{EqnAnalyticCellSystem}). When the region is bounded, the system (\ref{EqnAnalyticCellSystem}) of inequalities essentially forms a polyhedral cell defined as
\begin{equation}\label{EqnAnalyticCell}
\mathcal{C}_F^{\tilde{r}} = \{ \mathbf{x} \in \mathbb{R}^3 | (\mathbf{I} - 2\textrm{diag}(\mathbf{s}(\tilde{r})) \mathbf{A}^{\tilde{r}}\mathbf{x} \preceq 0 \} ,
\end{equation}
which we term as \emph{analytic cell of a SDF's zero-level isosurface}, shortened as \emph{analytic cell}. We note that an analytic cell could also be a region open towards infinity in some directions.

Given the plane functional (\ref{EqnSDFPlaneFunctional}), we define the polygonal face that is an intersection of analytic cell $\tilde{r}$ and surface $\mathcal{Z}$ as
\begin{equation}\label{EqnAnalyticFace}
\mathcal{P}_F^{\tilde{r}} = \{ \mathbf{x} \in \mathbb{R}^3 | \mathbf{a}_F^{\tilde{r}}\mathbf{x} = 0 , (\mathbf{I} - 2\textrm{diag}(\mathbf{s}(\tilde{r})) \mathbf{A}^{\tilde{r}}\mathbf{x} \preceq 0 \} ,
\end{equation}
which we term less precisely as \emph{analytic face of a SDF's zero-level isosurface}, shortened as \emph{analytic face}, since it is possible that the face goes towards infinity in some directions. With the analytic form (\ref{EqnAnalyticFace}), $\mathcal{Z}$ realized by a ReLU based MLP thus defines a piecewise planar surface, which could be an approximation to an underlying smooth surface when the SDF $F$ is trained using techniques presented shortly in Section \ref{SecSDFTrain}.

\subsection{A closed mesh via connected cells and faces}

We have not so far specified the types of surface that $\mathcal{Z} = \{ \mathbf{x} \in \mathbb{R}^3 | f\circ\mathbf{T}(\mathbf{x}) = 0 \}$ can represent, as long as they are piecewise planar whose associated analytic cells and faces respectively satisfy (\ref{EqnAnalyticCell}) and (\ref{EqnAnalyticFace}). In practice, one is mostly interested in those surface type representing the boundaries of non-degenerate 3D objects, which means that the objects do not have infinitely thin parts and a surface properly separates the interior and exterior of its object (cf. Figure 1.1 in \cite{MarioBotschPolygonalMesh} for an illustration). This type of object surface usually has the property of being continuous and closed.

A closed, piecewise planar surface $\mathcal{Z}$ means that every planar face of the surface is connected with other faces via shared edges. We have the following theorem that characterizes the conditions under which analytic faces (\ref{EqnAnalyticFace}) in their respective analytic cells (\ref{EqnAnalyticCell}) guarantee to connect and form a closed, piecewise planar surface.
\begin{theorem}\label{MainResult}
Assume that the zero-level isosurface $\mathcal{Z}$ of a SDF $F = f\circ\mathbf{T}$ defines a closed, piecewise planar surface. If for any region/cell $r\in \mathcal{R}(\mathbf{T})$, its associated linear mapping $\mathbf{T}^r$ (\ref{EqnMLPRegionMapping}) and the induced plane $\mathbf{a}_F^r = \mathbf{w}_f^{\top}\mathbf{T}^r$ (\ref{EqnSDFPlaneFunctional}) are uniquely defined, i.e., $\mathbf{T}^r \neq \beta\mathbf{T}^{r'}$ and $\mathbf{a}_F^r \neq \beta\mathbf{a}_F^{r'}$ for any region pair of $r$ and $r'$, where $\beta$ is an arbitrary scaling factor, then analytic faces $\{ \mathcal{P}_F^{\tilde{r}} \}$ defined by (\ref{EqnAnalyticFace}) connect and exactly form the surface $\mathcal{Z}$.
\end{theorem}
\vspace{-0.4cm}
\begin{proof}
The proof is given in Appendix \ref{AppendixMainResultProof}.
\end{proof}

We note that the conditions assumed in Theorem \ref{MainResult} are practically reasonable up to a numerical precision of the learned weights in the SDF $F = f\circ\mathbf{T}$. The proof of theorem also suggests an algorithm to identify the polygonal faces of a surface mesh learned by $F$, which is to be presented shortly.

\section{The proposed analytic marching algorithm}\label{SecAMAlgm}

Given a learned SDF $F = f\circ\mathbf{T}$ whose zero-level isofurface $\mathcal{Z} = \{ \mathbf{x} \in \mathbb{R}^3 | F(\mathbf{x}) = 0 \}$ defines a closed, piecewise planar surface, Theorem \ref{MainResult} suggests that obtaining the mesh of $\mathcal{Z}$ concerns with identification of analytic faces $\mathcal{P}_F^{\tilde{r}}$ in $\{\mathcal{C}_F^{\tilde{r}} | \tilde{r} \in \widetilde{\mathcal{R}} \}$. To this end, we propose an algorithm of \emph{analytic marching} that marches among $\{\mathcal{C}_F^{\tilde{r}} | \tilde{r} \in \widetilde{\mathcal{R}} \}$ to identify vertices and edges of the polygonal faces $\{\mathcal{P}_F^{\tilde{r}} | \tilde{r} \in \widetilde{\mathcal{R}} \}$, where the name is indeed to show respect to the classical discrete algorithm of marching cubes \cite{MC}.

Specifically, analytic marching is triggered by identifying at least one point $\mathbf{x} \in \mathcal{Z}$. Given the parametric model $F$ and an arbitrarily initialized point $\mathbf{x} \in \mathbb{R}^3$, this can be simply achieved by solving the following problem with stochastic gradient descent (SGD)
\begin{equation}\label{EqnSurfacePointOptim}
\min_{\mathbf{x} \in \mathbb{R}^3} | F(\mathbf{x}) | .
\end{equation}
For a point $\mathbf{x}$ with $F(\mathbf{x}) = 0$, its state vector $\mathbf{s}(\mathbf{x})$ can be computed via (\ref{EqnMLPStateFunctional}), which specifies the analytic cell $\mathcal{C}_F^{\tilde{r}_{\mathbf{x}}}$ (\ref{EqnAnalyticCell}) and analytic face $\mathcal{P}_F^{\tilde{r}_{\mathbf{x}}}$ (\ref{EqnAnalyticFace}) where $\mathbf{x}$ resides. Initialize an active set $\mathcal{S}^{\bullet} = \emptyset$ and an inactive set $\mathcal{S}^{\circ} = \emptyset$. Push $\mathbf{s}(\mathbf{x})$ into $\mathcal{S}^{\bullet}$. Analytic marching proceeds by repeating the following steps.
\begin{enumerate}
\item Take an active state $\mathbf{s}_i$ from $\mathcal{S}^{\bullet}$, which specifies its analytic cell $\mathcal{C}_F^{\tilde{r}_i}$ and analytic face $\mathcal{P}_F^{\tilde{r}_i}$.

\item To compute the set $\mathcal{V}_{\mathcal{P}}^{\tilde{r}_i}$ of vertex points associated with $\mathcal{P}_F^{\tilde{r}_i}$, enumerates all the pair $(H_{lk}^{\tilde{r}_i}, H_{l'k'}^{\tilde{r}_i})$ of boundary planes $\{ H_{lk}^{\tilde{r}_i}\}$ defined by (\ref{EqnBoundaryPlanes}), with $l = 1, \dots, L$ and $k = 1, \dots, n_l$. Each pair $(H_{lk}^{\tilde{r}_i}, H_{l'k'}^{\tilde{r}_i})$, together with $\mathcal{P}_F^{\tilde{r}_i}$, defines the following system of three equations
    \begin{equation}\label{EqnVertexEquationSystem}
    \mathbf{B}^{\tilde{r}_i}\mathbf{x} = 0 ,
    \end{equation}
    where $\mathbf{B}^{\tilde{r}_i} = [\mathbf{a}_{lk}^{\tilde{r}_i}; \mathbf{a}_{l'k'}^{\tilde{r}_i}; \mathbf{a}_F^{\tilde{r}_i}] \in \mathbb{R}^{3\times 3}$.

\item System (\ref{EqnVertexEquationSystem}) gives a potential vertex $\mathbf{v}$ corresponding to the boundary pair $(H_{lk}^{\tilde{r}_i}, H_{l'k'}^{\tilde{r}_i})$. Validity of $\mathbf{v}$ is checked by the boundary condition (\ref{EqnAnalyticCellSystem}) of the cell $\mathcal{C}_F^{\tilde{r}_i}$; if it is true, we have a vertex $\mathbf{v} \in \mathcal{V}_{\mathcal{P}}^{\tilde{r}_i}$. All the valid vertices obtained by solving (\ref{EqnVertexEquationSystem}) form $\mathcal{V}_{\mathcal{P}}^{\tilde{r}_i}$ of the face $\mathcal{P}_F^{\tilde{r}_i}$, whose pair-wise edges are defined by those on the same boundary planes.

\item Record all the boundary planes $\{ \widehat{H}_{lk}^{\tilde{r}_i} \}$ of $\mathcal{C}_F^{\tilde{r}_i}$ that give valid vertices. Proof of Theorem \ref{MainResult} tells that the surface $\mathcal{Z}$ is in general well positioned in $\{\mathcal{C}_F^{\tilde{r}} | \tilde{r} \in \widetilde{\mathcal{R}} \}$ (cf. proof of Theorem \ref{MainResult} for details), and the analytic cell connecting $\mathcal{C}_F^{\tilde{r}_i}$ at a boundary plane $\widehat{H}_{lk}^{\tilde{r}_i}$ has its state vector switching only at the $k^{th}$ neuron of layer $l$, which gives a new state $\hat{\mathbf{s}}_i$ and thus a new analytic cell.

\item Push $\mathbf{s}_i$ out of the active set $\mathcal{S}^{\bullet}$ and into the inactive set $\mathcal{S}^{\circ}$. Push $\{\hat{\mathbf{s}}_i | \hat{\mathbf{s}}_i \not\in  \mathcal{S}^{\circ} \}$ into the active set $\mathcal{S}^{\bullet}$.
\end{enumerate}
The algorithm of analytic marching proceeds by repeating the above steps, until the active set $\mathcal{S}^{\bullet}$ is cleared up.

\noindent\textbf{Algorithmic guarantee} Theorem \ref{MainResult} guarantees that when the SDF $F$ learns its zero-level isosurface $Z$ as a closed, piecewise planar surface, identification of all the analytic faces forms the closed surface mesh. The proposed analytic marching algorithm is anchored at the cell state transition of the above step 4, whose success is of high probability due to a phenomenon similar to the blessing of dimensionality \cite{DimBlessing}. More specifically, it is of low probability that edges connecting planar faces of $\mathcal{Z}$ coincide with those of analytic cells (cf. proof of Theorem \ref{MainResult} for detailed analysis). 

\subsection{Analysis of computational complexities}\label{SecComplexityAnalysis}

Assume that the SDF $F = f\circ \mathbf{T}$ is built on an MLP of $L$ hidden layers, each of which has $n_l$ neurons, $l = 1, \dots, L$. Let $N = n_1 + \dots, n_L$. For ease of analysis, we assume $n_1 = \cdots = n_L = n$ and thus $N = nL$. The computations inside each analytic cell concern with computing the boundary planes, solving a maximal number of ${N}\choose{2}$ equation systems (\ref{EqnVertexEquationSystem}), and checking the validity of resulting vertices, which give a complexity order of $\mathcal{O}(n^3L^3)$. Improving step 2 of the algorithm with pivoting operation \cite{pivot} avoids enumeration of all pairs of boundary planes, reducing the complexity to an order of $\mathcal{O}(|\mathcal{V_{\mathcal{P}}}|n^2L^2)$, where $|\mathcal{V_{\mathcal{P}}}|$ represents the number of vertices per face and is typically less than 10.
We know from \cite{Montufar2014} that the maximal size of the set $\mathcal{R}(\mathbf{T})$ of linear regions in general has an order of $\mathcal{O}\left( (n/n_0)^{(L-1)n_0} n^{n_0}  \right)$, where $n_0$ is the dimensionality of input space. Since our focus of interest is the 2-dimensional object surface embedded in the 3D space, we have $n_0 = 2$ and thus the maximal size of $\mathcal{R}(\mathbf{T})$, which bounds the maximal number of analytic cells, in general has an order of $\mathcal{O}\left( (n/2)^{2(L-1)} n^2  \right)$. Overall, the complexity of our analytic marching algorithm has an order of $\mathcal{O}\left( (n/2)^{2(L-1)} |\mathcal{V_{\mathcal{P}}}|n^4L^2 \right)$, which is exponential w.r.t. the MLP depth $L$ and polynomial w.r.t. the MLP width $n$.

The above analysis shows that the complexity nature of our algorithm is the complexity of SDF function, which is completely different from those of existing algorithms, such as marching cubes \cite{MC}, whose complexities are irrelevant to function complexities but rather depend on the discretized resolutions of the 3D space. Our algorithm thus provides an opportunity to recover highly precise mesh reconstruction by learning MLPs of low complexities. Alternatively, one may resort to network compression/pruning techniques \cite{HanNIPS15,LuoICCV17}, which can potentially reduce network complexities with little sacrifice of inference precision.

\subsection{Practical implementations and parallel efficiency}\label{SecPracticalParallelImplementation}

The proposed analytic marching can be naturally implemented in parallel. Instead of triggering the algorithm from a single $\mathbf{x}$ by solving (\ref{EqnSurfacePointOptim}), we can practically initialize as many as possible such points from the 3D space, and the algorithm would proceed in parallel. The parallel implementation can also be enhanced by simultaneously marching towards all the neighboring cells of the current one (cf. steps 4 and 5 of the algorithm). Since the SDF $F = f\circ \mathbf{T}$ is learned from training samples of ground-truth object mesh, \emph{its zero-level isosurface is practically not guaranteed to be exactly the same as the ground truth, and in many cases, it is not even closed}; consequently, the condition assumed in Theorem \ref{MainResult} is not satisfied. In such cases, initialization of multiple surface points would help recover components of the surface that are possibly isolated, whose efficacy is verified in Section \ref{SecExps}.

\section{Training of deep implicit surface networks}\label{SecSDFTrain}

For any $\mathbf{x} \in \mathbb{R}^3$, let $d(\mathbf{x})$ denote its ground-truth value of signed distance to the surface. We use the following regularized objective to train a SDF $F = f\circ \mathbf{T}$,
\begin{equation}\label{EqnSDFTrainObj}
\min_{F = f\circ \mathbf{T}} \mathbb{E}_{\mathbf{x} \sim \mathbb{R}^3} \big| F(\mathbf{x}) - d(\mathbf{x}) \big| + \alpha \big| \| \partial F(\mathbf{x}) / \partial\mathbf{x} \| - 1 \big| ,
\end{equation}
where $\alpha$ is a penalty parameter. The unit gradient regularizer follows \cite{Michalkiewicz2019}, which aims to promote learning of a smooth gradient field.

\section{Experiments}\label{SecExps}

\noindent\textbf{Datasets}
We use five categories of ``Rifle'', ``Chair'', ``Airplane'', ``Sofa'', and ``Table'' from the ShapeNetCoreV1 dataset \cite{ShapeNet}, 200 object instances per category, for evaluation of different meshing algorithms. The 3D space containing mesh models of these instances is normalized in $[-1, 1]^3$. To train an MLP based SDF, we follow \cite{DISN} and sample more points in the 3D space that are in the vicinity of the surface. Ground-truth SDF values are calculated by linear interpolation from a dense SDF grid obtained by \cite{GenerateSDF1,GenerateSDF2}.

\noindent\textbf{Implementation details}
Our training hyperparameters are as follows. The learning rates start at 1e-3, and decay every 20 epoches by a factor of 10, until the total number of 60 epoches. We set weight decay as 1e-4 and the penalty in (\ref{EqnSDFTrainObj}) as $\alpha = 0.01$. In all experiments, we trigger our algorithm from 100 randomly sampled points. As described in Section \ref{SecPracticalParallelImplementation}, our algorithm naturally supports parallel implementation, which however, has not been customized so far; \emph{the current algorithm is simply implemented on a CPU (Intel E5-2630 @ 2.20GHz) in a straightforward manner. For comparative algorithms such as marching cubes \cite{MC}, their dominating computations of evaluating SDF values of sampled discrete points are implemented on a GPU (Tesla K80)}, which certainly gives them an unfair advantage. Nevertheless, we show in the following a better trade-off of precision and efficiency from our algorithm, even under the unfair comparison.

\noindent\textbf{Evaluation metrics}
We use the following metrics to quantitatively measure the accuracies between recovered mesh results and ground-truth ones: (1) Chamfer Distance (CD), (2) Earth Mover’s Distance (EMD), (3) Intersection over Union (IoU), and (4) F-score (F), where poisson disk sampling \cite{PoissonSampling} is used to sample points from surface. For the measures of IoU and F-Score, the larger the better, and for CD and EMD, the smaller the better. These metrics provide complementary perspectives to compare different algorithms. Additionally, wall-clock time and number of faces in the recovered meshes are reported as reference.

\subsection{Ablation studies}

Analysis in Section \ref{SecAMAlgm} tells that our proposed analytic marching is possible to exactly recover the mesh captured by a learned MLP. \emph{We note that the zero-level isosurface of a learned MLP only approximates the ground-truth mesh}, and the approximation quality mostly depends on the capacity of the MLP network, which is in turn determined by network depth and network width. We study in this section how the network depth and width affect the recovery accuracies and efficiency of our proposed algorithm.

We conduct experiments by fixing two groups respectively of the same numbers of MLP neurons, while varying either the numbers of layers or the numbers of neurons per layer. The first group uses a total of $360$ neurons, whose depth/width distributions are D4-W90, D6-W60, and D8-W45, where ``D'' stands for depth and ``W'' stands for width. The second group uses a total of $900$ neurons, whose depth/width distributions are D10-W90, D15-W60, and D20-W45. Results in Table \ref{TableAblation} tell that under different evaluation metrics, accuracies of the recovered meshes consistently improve with the increased network capacities, and the numbers of mesh faces and inference time are increased as well. Given a fixed number of neurons, it seems that properly deep networks are advantageous in terms of precision-efficiency trade off. Since the experimental settings fall in the (relatively) shallower range, polynomial increase of network width dominates the computational complexity, as analyzed in Section \ref{SecComplexityAnalysis}.

Results in Table \ref{TableAblation} are from the experiments on the 200 instances of ``Rifle'' category. Results of other object categories are of similar quality. We summarize in Table \ref{TableAllCategories} results of all the five categories based on an MLP of 6 layers with 60 neurons per layer (the D6-W60 setting in Table \ref{TableAblation}), which tell that the surface and/or topology complexities of ``Chair'' are higher, and those of ``Airplane'' are lower. 

\begin{table*}[!htbp]
\caption{Ablation studies by varying the numbers of layers and the numbers of neurons per layer for the trained MLPs of SDF. ``D'' stands for network depth and ``W'' for width. Results are from the 200 instances of ``Rifle'' categories. F-scores use $\tau = 5\times10^{-3}$. } 
\label{TableAblation}
\vskip 0.1in
\begin{center}
\begin{small}
\begin{tabular}{cccccccc}
\toprule
Architecture & CD($\times10^{-1}$) & EMD($\times10^{-3}$) & IoU($\%$) & F@$\tau$($\%$) & F@$2\tau$($\%$) & Face No. & Time(sec.) \\
\midrule
D4-W90 & 6.16 & 8.60 & 86.5 & 84.8 & 95.4 & 97865 & 17.0 \\
D6-W60 & 5.29 & 8.00 & 86.9 & 85.5 & 95.8 & 100179 & 15.0 \\
D8-W45 & 5.67 & 8.12 & 85.7 & 84.2 & 95.4 & 93482 & 9.75 \\
\midrule
D10-W90 & 3.64 & 5.85 & 90.2 & 88.1 & 98.6 & 421840 & 173 \\
D15-W60 & 3.85 & 6.23 & 88.9 & 87.4 & 97.5 & 336044 & 86.8 \\
D20-W45 & 4.21 & 6.89 & 86.1 & 86.6 & 95.5 & 253970 & 47.3 \\

\bottomrule
\end{tabular}
\end{small}
\end{center}
\vskip -0.1in
\end{table*}

\begin{table*}[!htbp]
\caption{Results of all the five categories based on an MLP of 6 layers with 60 neurons per layer (the D6-W60 setting in Table \ref{TableAblation}). F-scores use $\tau = 5\times10^{-3}$.} 
\label{TableAllCategories}
\vskip 0.1in
\begin{center}
\begin{small}
\begin{tabular}{cccccccc}
\toprule
Category & CD($\times10^{-1}$) & EMD($\times10^{-3}$) & IoU($\%$) & F@$\tau$($\%$) & F@$2\tau$($\%$) & Face No. & Time(sec.) \\
\midrule
Rifle & 5.29 & 8.00 & 86.9 & 85.5 & 95.8 & 100179 & 15.0 \\
Chair & 7.27 & 8.01 & 89.6 & 52.9 & 96.1 & 182624 & 25.3 \\
Airplane & 3.25 & 5.32 & 89.4 & 89.8 & 97.9 & 131074 & 19.0 \\
Sofa & 6.78 & 6.65 & 96.6 & 53.2 & 97.9 & 156863 & 22.0 \\
Table & 6.98 & 7.16 & 90.7 & 51.6 & 96.8 & 163395 & 22.7 \\
Mean & 5.91 & 7.03 & 90.6 & 66.6 & 96.9 & 146827 & 20.8 \\
\bottomrule
\end{tabular}
\end{small}
\end{center}
\vskip -0.1in
\end{table*}

\subsection{Comparisons with existing meshing algorithms}

In this section, we compare our proposed analytic meshing (AM) with existing algorithms of greedy meshing (GM), marching cubes (MC) \cite{MC}, marching tetrahedra (MT) \cite{MT}, and dual contouring (DC) \cite{DC}, where MC is the de-facto standard meshing solution adopted by many surface meshing applications, and DC improves over MC with ingredients including partitioning the 3D space with a hierarchical structure of octree. These comparative algorithms are all based on discretizing the 3D space by evaluating the SDF values at a regular grid of sampled points; consequently, their meshing accuracies and efficiency depend on the sampling resolution. We thus implement them under a range of sampling resolutions from $32^3$ to a GPU memory limit of $512^3$.

\begin{figure*}[!htbp]
    \vskip 0.1in
    \begin{center}
    \includegraphics[scale=0.5]{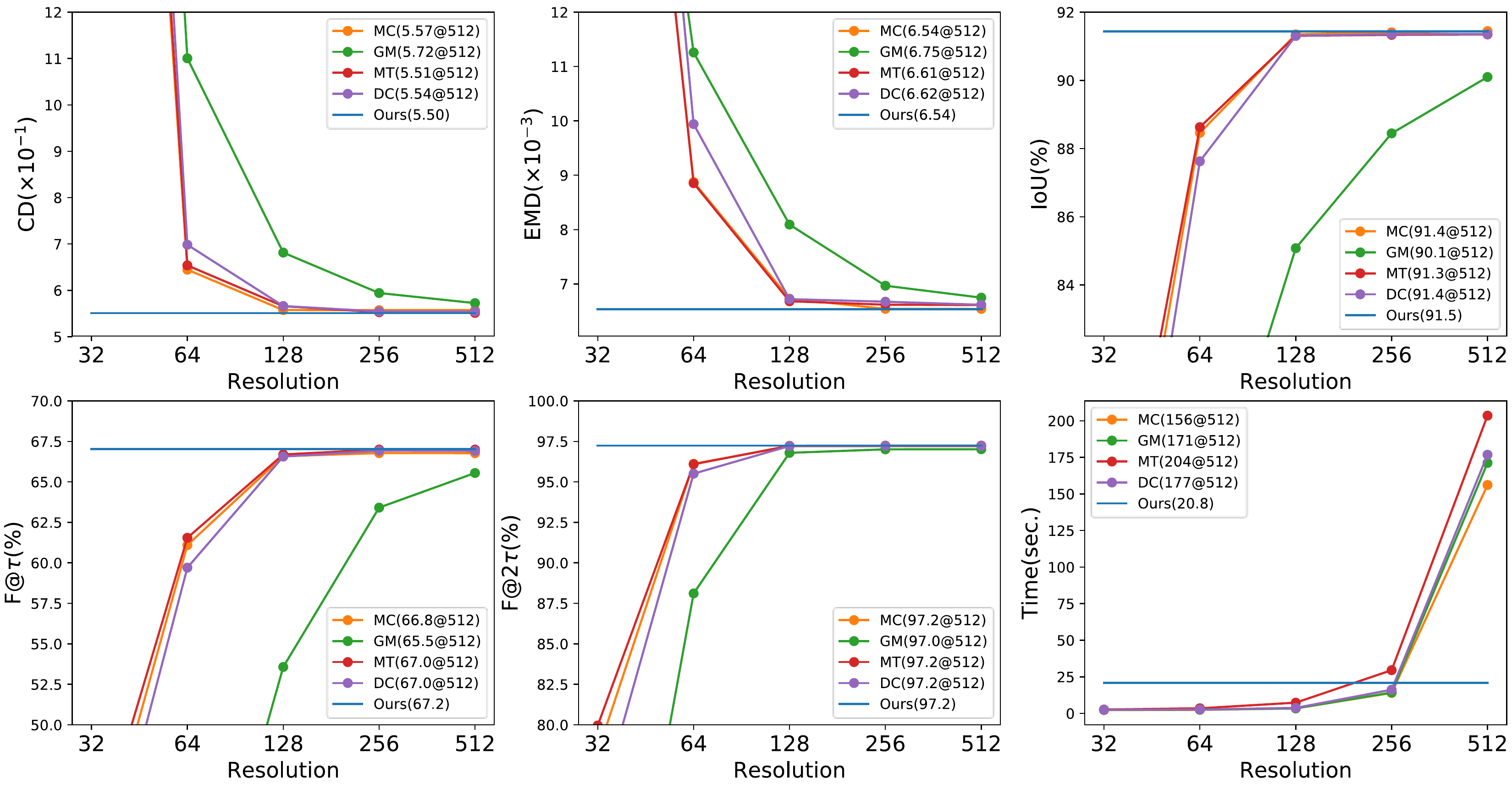}
    \caption{Quantitative comparisons of different meshing algorithms under metrics of recovery precision and inference world-clock time. For greedy meshing (GM), marching cubes (MC), marching tetrahedra (MT), and dual contouring (DC), results under a resolution range of discrete point sampling from $32^3$ to a GPU memory limit of $512^3$ are presented, and the dominating computations of their sampled points' SDF values are implemented on GPU. Numerical results of this figure are given in Appendix \ref{AppendixNumericalComp}. }
    \label{FigQuantitativeCurves}
    \end{center}
    \vskip -0.1in
\end{figure*}

Figure \ref{FigQuantitativeCurves} shows that among these comparative methods, marching cubes in general performs better in terms of a balanced precision and efficiency. However, under different evaluation metrics, recovery accuracies of these methods are upper bounded by our proposed one. As noted in the implementation details of this section, the dominating computations of these methods are implemented on GPU, which gives them an unfair advantage of computational efficiency. Nevertheless, results in Figure \ref{FigQuantitativeCurves} tell that even under the unfair comparison, our algorithm is much faster at a similar level of recovery precision.

\begin{figure*}[!htbp]
    \vskip 0.1in
    \begin{center}
    \includegraphics[scale=0.22]{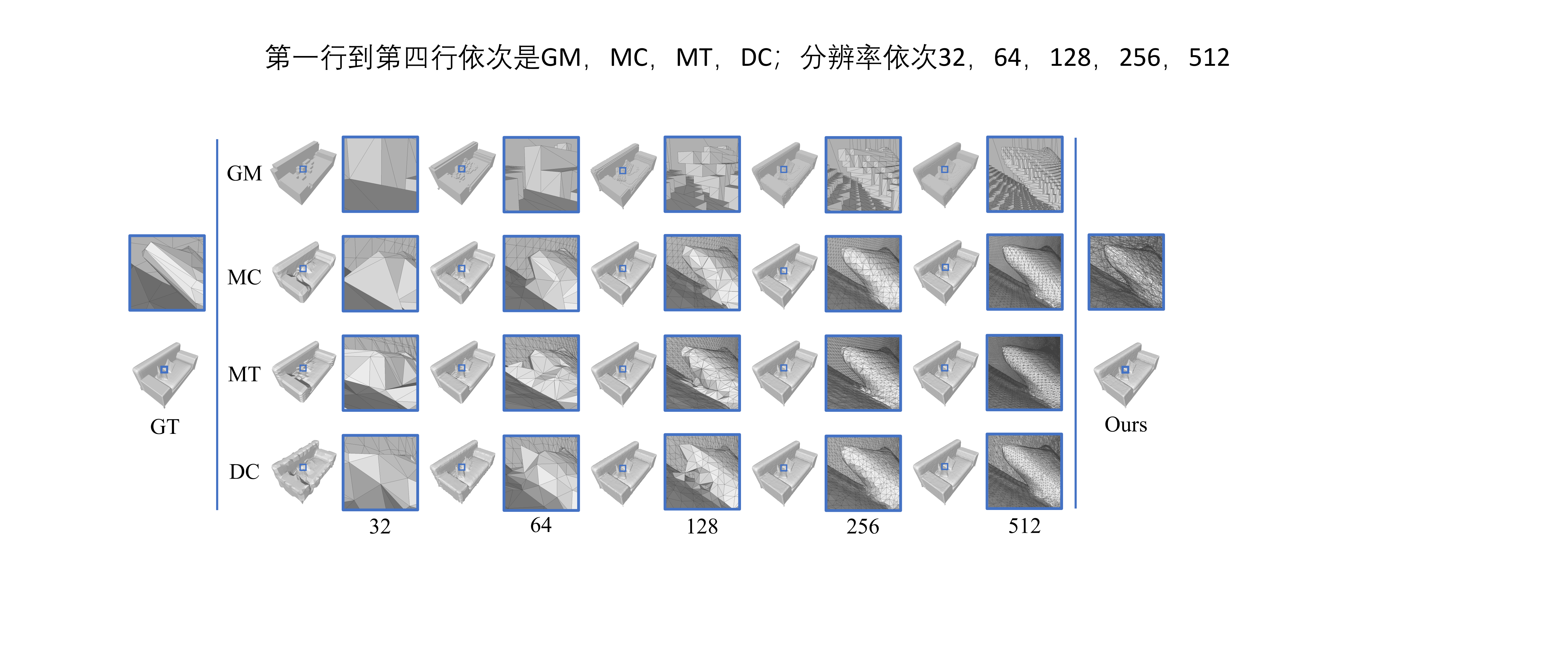}
    \caption{Qualitative comparisons of different meshing algorithms. For greedy meshing (GM), marching cubes (MC), marching tetrahedra (MT), and dual contouring (DC), results under a resolution range of discrete point sampling from $32^3$ to a GPU memory limit of $512^3$ are presented. }
    \label{FigQualityResults}
    \end{center}
    \vskip -0.1in
\end{figure*}

The quantitative results in Figure \ref{FigQuantitativeCurves} are averaged ones over all the object instances of all the five categories, using an MLP of 6 layers with 60 neurons per layer (the D6-W60 setting in Table \ref{TableAblation}). We finally show qualitative results in Figure \ref{FigQualityResults}, where mesh results of an example object are presented. More qualitative results are given in Appendix \ref{AppendixAdditionalQualityResults}. Our proposed algorithm is particularly advantageous in capturing geometry details on high-curvature surface areas.

\section{Conclusion}

In this work, we contribute an analytic meshing solution from learned deep implicit surface networks. Our contribution is motivated by the established knowledge that a ReLU based MLP partitions its input space into a number of linear regions. We identify from these regions analytic cells and analytic faces that are associated with the zero-level isosurface of the learned MLP based implicit function. We prove that under mild conditions, the identified analytic faces are guaranteed to connect and form a closed, piecewise planar surface. Our theorem inspires us to propose a naturally parallelizable algorithm of analytic marching, which marches among analytic cells to exactly recover the mesh captured by a learned MLP. Experiments on benchmark dataset of 3D object repositories confirm the advantages of our algorithm over existing ones.


\nocite{langley00}

\bibliography{PAPER}

\begin{thebibliography}{29}
\providecommand{\natexlab}[1]{#1}
\providecommand{\url}[1]{\texttt{#1}}
\expandafter\ifx\csname urlstyle\endcsname\relax
  \providecommand{\doi}[1]{doi: #1}\else
  \providecommand{\doi}{doi: \begingroup \urlstyle{rm}\Url}\fi

\bibitem[Avis \& Fukuda(1991)Avis and Fukuda]{pivot}
Avis, D. and Fukuda, K.
\newblock A pivoting algorithm for convex hulls and vertex enumeration of
  arrangements and polyhedra.
\newblock volume~8, pp.\  98--104, 01 1991.

\bibitem[Blinn(1982)]{blobby_molecules}
Blinn, J.~F.
\newblock A generalization of algebraic surface drawing.
\newblock \emph{ACM Trans. Graph.}, 1\penalty0 (3):\penalty0 235–256, July
  1982.

\bibitem[Botsch et~al.(2010)Botsch, Kobbelt, Pauly, Alliez, and
  Levy]{MarioBotschPolygonalMesh}
Botsch, M., Kobbelt, L., Pauly, M., Alliez, P., and Levy, B.
\newblock \emph{Polygon Mesh Processing}.
\newblock CRC Press, 2010.

\bibitem[Bowers et~al.(2010)Bowers, Wang, Wei, and Maletz]{PoissonSampling}
Bowers, J., Wang, R., Wei, L.-Y., and Maletz, D.
\newblock Parallel poisson disk sampling with spectrum analysis on surfaces.
\newblock In \emph{ACM SIGGRAPH Asia}, 2010.

\bibitem[Carr et~al.(1997)Carr, Beatson, and Fright]{RBF2}
Carr, J., Beatson, R., and Fright, W.
\newblock Surface interpolation with radial basis functions for medical
  imaging.
\newblock \emph{IEEE Transactions on Medical Imaging}, 16\penalty0 (1), 2 1997.

\bibitem[Carr et~al.(2001)Carr, Beatson, Cherrie, Mitchell, Fright, McCallum,
  and Evans]{RBF1}
Carr, J.~C., Beatson, R.~K., Cherrie, J.~B., Mitchell, T.~J., Fright, W.~R.,
  McCallum, B.~C., and Evans, T.~R.
\newblock Reconstruction and representation of 3d objects with radial basis
  functions.
\newblock In \emph{Proceedings of the 28th Annual Conference on Computer
  Graphics and Interactive Techniques}, pp.\  67–76, 2001.

\bibitem[{Chang} et~al.(2015){Chang}, {Funkhouser}, {Guibas}, {Hanrahan},
  {Huang}, {Li}, {Savarese}, {Savva}, {Song}, {Su}, {Xiao}, {Yi}, and
  {Yu}]{ShapeNet}
{Chang}, A.~X., {Funkhouser}, T., {Guibas}, L., {Hanrahan}, P., {Huang}, Q.,
  {Li}, Z., {Savarese}, S., {Savva}, M., {Song}, S., {Su}, H., {Xiao}, J.,
  {Yi}, L., and {Yu}, F.
\newblock {ShapeNet: An Information-Rich 3D Model Repository}.
\newblock \emph{arXiv:1512.03012}, 2015.

\bibitem[Chu et~al.(2018)Chu, Hu, Hu, Wang, and Pei]{OpenBox}
Chu, L., Hu, X., Hu, J., Wang, L., and Pei, J.
\newblock Exact and consistent interpretation for piecewise linear neural
  networks: {A} closed form solution.
\newblock In \emph{Proceedings of the 24th {ACM} {SIGKDD} International
  Conference on Knowledge Discovery {\&} Data Mining}, pp.\  1244--1253, 2018.

\bibitem[Curless \& Levoy(1996)Curless and Levoy]{TSDFCurlessL96}
Curless, B. and Levoy, M.
\newblock A volumetric method for building complex models from range images.
\newblock In \emph{SIGGRAPH}, pp.\  303--312. ACM, 1996.

\bibitem[Doi \& Koide(1991)Doi and Koide]{MT}
Doi, A. and Koide, A.
\newblock An efficient method of triangulating equivalued surfaces by using
  tetrahedral cells.
\newblock \emph{IEICE Transactions on Information and Systems}, 74, 01 1991.

\bibitem[Glorot et~al.(2011)Glorot, Bordes, and Bengio]{ReLU}
Glorot, X., Bordes, A., and Bengio, Y.
\newblock Deep sparse rectifier neural networks.
\newblock In \emph{Proceedings of the International Conference on Artificial
  Intelligence and Statistics}, 2011.

\bibitem[Gorban \& Tyukin(2018)Gorban and Tyukin]{DimBlessing}
Gorban, A. and Tyukin, I.
\newblock Blessing of dimensionality: mathematical foundations of the
  statistical physics of data.
\newblock \emph{arXiv:1801.03421}, 2018.

\bibitem[Han et~al.(2015)Han, Pool, Tran, and Dally]{HanNIPS15}
Han, S., Pool, J., Tran, J., and Dally, W.
\newblock Learning both weights and connections for efficient neural network.
\newblock In \emph{Advances in neural information processing systems}, pp.\
  1135--1143, 2015.

\bibitem[{Jia} et~al.(2019){Jia}, {Li}, {Wen}, {Liu}, and {Tao}]{OrthDNNs}
{Jia}, K., {Li}, S., {Wen}, Y., {Liu}, T., and {Tao}, D.
\newblock Orthogonal deep neural networks.
\newblock \emph{IEEE Transactions on Pattner Analysis and Machine
  Intelligence}, 2019.

\bibitem[Ju et~al.(2002)Ju, Losasso, Schaefer, and Warren]{DC}
Ju, T., Losasso, F., Schaefer, S., and Warren, J.
\newblock Dual contouring of hermite data.
\newblock \emph{ACM Trans. Graph.}, 21\penalty0 (3):\penalty0 339–346, July
  2002.

\bibitem[Lorensen \& Cline(1987)Lorensen and Cline]{MC}
Lorensen, W.~E. and Cline, H.~E.
\newblock Marching cubes: A high resolution 3d surface construction algorithm.
\newblock In \emph{SIGGRAPH}, pp.\  163--169. ACM, 1987.

\bibitem[Luo et~al.(2017)Luo, Wu, and Lin]{LuoICCV17}
Luo, J.-H., Wu, J., and Lin, W.
\newblock Thinet: A filter level pruning method for deep neural network
  compression.
\newblock In \emph{IEEE international conference on computer vision}, pp.\
  5058--5066, 2017.

\bibitem[Michalkiewicz et~al.(2019)Michalkiewicz, Pontes, Jack, Baktashmotlagh,
  and Eriksson]{Michalkiewicz2019}
Michalkiewicz, M., Pontes, J.~K., Jack, D., Baktashmotlagh, M., and Eriksson,
  A.
\newblock Deep level sets: Implicit surface representations for 3d shape
  inference.
\newblock \emph{arXiv:1901.06802}, 2019.

\bibitem[Mont\'{u}far et~al.(2014)Mont\'{u}far, Pascanu, Cho, and
  Bengio]{Montufar2014}
Mont\'{u}far, G., Pascanu, R., Cho, K., and Bengio, Y.
\newblock On the number of linear regions of deep neural networks.
\newblock In \emph{Advances in Neural Information Processing Systems}, pp.\
  2924--2932, 2014.

\bibitem[Nishimura et~al.(1985)Nishimura, Hirai, Kawai, Kawata, Shirkawa, and
  Omura]{metaballs}
Nishimura, H., Hirai, M., Kawai, T., Kawata, T., Shirkawa, I., and Omura, K.
\newblock Object modeling by distributed function and a method of image
  generation (in japanese).
\newblock \emph{The Transactions of the Institute of Electronics and
  Communication Engineers of Japan}, 68, 01 1985.

\bibitem[Orlik \& Terao(1992)Orlik and Terao]{orlik1992arrangements}
Orlik, P. and Terao, H.
\newblock \emph{Arrangements of Hyperplanes}.
\newblock Springer Berlin Heidelberg, 1992.

\bibitem[Park et~al.(2019)Park, Florence, Straub, Newcombe, and
  Lovegrove]{DeepSDF}
Park, J.~J., Florence, P., Straub, J., Newcombe, R., and Lovegrove, S.
\newblock Deepsdf: Learning continuous signed distance functions for shape
  representation.
\newblock \emph{arXiv:1901.05103}, 2019.

\bibitem[Pascanu et~al.(2014)Pascanu, Mont{\'u}far, and
  Bengio]{pascanu2013number}
Pascanu, R., Mont{\'u}far, G., and Bengio, Y.
\newblock On the number of response regions of deep feed forward networks with
  piece-wise linear activations.
\newblock In \emph{International Conference on Learning Representations}, 2014.

\bibitem[Sin et~al.(2013)Sin, Schroeder, and Barbič]{GenerateSDF1}
Sin, F., Schroeder, D., and Barbič, J.
\newblock Vega: Non‐linear fem deformable object simulator.
\newblock \emph{Computer Graphics Forum}, 32, 2013.

\bibitem[Turk \& O’Brien(1999)Turk and O’Brien]{RBF3}
Turk, G. and O’Brien, J.~F.
\newblock Shape transformation using variational implicit functions.
\newblock In \emph{Proceedings of the 26th Annual Conference on Computer
  Graphics and Interactive Techniques}, pp.\  335–342, 1999.

\bibitem[Wyvill et~al.(1986)Wyvill, McPheeters, and Wyvill]{soft_object}
Wyvill, G., McPheeters, C., and Wyvill, B.
\newblock Data structure for soft objects.
\newblock \emph{The Visual Computer}, 2:\penalty0 227--234, 08 1986.

\bibitem[Xu \& Barbič(2014)Xu and Barbič]{GenerateSDF2}
Xu, H. and Barbič, J.
\newblock Signed distance fields for polygon soup meshes.
\newblock \emph{Proceedings - Graphics Interface}, pp.\  35--41, 01 2014.

\bibitem[Xu et~al.(2019)Xu, Wang, Ceylan, Mech, and Neumann]{DISN}
Xu, Q., Wang, W., Ceylan, D., Mech, R., and Neumann, U.
\newblock Disn: Deep implicit surface network for high-quality single-view 3d
  reconstruction.
\newblock \emph{arXiv:1905.10711}, 2019.

\bibitem[Zaslavsky(1975)]{Zaslavsky1975}
Zaslavsky, T.
\newblock Facing up to arrangements: Face-count formulas for partitions of
  space by hyperplanes.
\newblock \emph{Number 154 in Memoirs of the American Mathematical Society},
  1975.

\end{thebibliography}
\bibliographystyle{icml2020}

\appendix

\section{Proof of Theorem \ref{MainResult}}\label{AppendixMainResultProof}

\begin{proof}
The proof proceeds by first showing that each planar face on the surface $\mathcal{Z}$ captured by the SDF $F = f\circ\mathbf{T}$ uniquely corresponds to an analytic face of an analytic cell, and then showing that for any pair of planar faces connected on $\mathcal{Z}$, their corresponding analytic faces are connected via boundaries of their respective analytic cells.

Let $\mathcal{P}_1$ denote a planar face on the surface $\mathcal{Z}$ 
, and $\mathbf{n}_1 \in \mathbb{R}^3$ be its normal. We have $\mathbf{n}_1^{\top}\mathbf{x} = 0 \ \forall \ \mathbf{x}\in \mathcal{P}_1$. Equation (\ref{EqnSDFPlaneFunctional}) tells that $\mathbf{n}_1$ must be proportional at least to one of $\{\mathbf{a}_F^r | r\in\mathcal{R} \}$. By the unique plane condition, i.e., each of $\{\mathbf{a}_F^r | r\in\mathcal{R} \}$ is uniquely defined, we have $\mathbf{n}_1 \propto \mathbf{a}_F^{r_1\top}$ of a certain region $r_1$. Assume $r_1$ is not an analytic cell, which suggests that there exists no intersection between $\mathbf{a}_F^{r_1}$ and $r_1$ and we have $\mathbf{a}_F^{r_1}\mathbf{x} = \mathbf{n}_1^{\top}\mathbf{x} \neq 0$ for all $\mathbf{x} \in r_1$, and thus $\mathcal{P}_1 \land r_1 = \emptyset$; it suggests that the normal $\mathbf{n}_1$ of $\mathcal{P}_1$ is induced in a different region $r_1'$ by $\mathbf{n}_1 \propto \mathbf{a}_F^{r_1'} = \mathbf{w}_f^{\top}\mathbf{T}^{r_1'}$, which contradicts with the assumed unique plane condition. We thus have that $r_1$ must be an analytic cell.

Let $\mathbf{n}_1 \propto \mathbf{a}_F^{\tilde{r}_1\top}$ of a certain analytic cell $\tilde{r}_1 \in \widetilde{\mathcal{R}}$ (or $\mathcal{C}_F^{\tilde{r}_1}$), and we have the analytic face $\mathcal{P}_F^{\tilde{r}_1} \subseteq \mathcal{P}_1$. Assume there exist $\mathcal{P}_1 - \mathcal{P}_F^{\tilde{r}_1} = \{ \mathbf{x} \in \mathcal{Z} | \mathbf{x} \in \mathcal{P}_1, \mathbf{x} \notin \mathcal{P}_F^{\tilde{r}_1} \}$, which means that for any $\mathbf{x} \in \mathcal{P}_1 - \mathcal{P}_F^{\tilde{r}_1}$, it resides in an analytic face $\mathcal{P}_F^{\tilde{r}_1'}$ of a different cell $\mathcal{C}_F^{\tilde{r}_1'}$; since $\mathbf{x} \in \mathcal{P}_1$, we have $\mathbf{n}_1 \propto \mathbf{a}_F^{\tilde{r}_1'\top}$ and thus $\mathbf{a}_F^{\tilde{r}_1} \propto \mathbf{a}_F^{\tilde{r}_1'}$, which contradicts with the unique plane condition of $\mathbf{a}_F^{\tilde{r}_1} \not\propto \mathbf{a}_F^{\tilde{r}_1'}$.
We thus have $\mathcal{P}_1 = \mathcal{P}_F^{\tilde{r}_1}$ and $\mathcal{P}_1 \subset \mathcal{C}_F^{\tilde{r}_1}$. By the definition (\ref{EqnAnalyticFace}) of analytic face, the above argument also tells that planar faces on $\mathcal{Z}$ and analytic faces $\{ \mathcal{C}_F^{\tilde{r}} | \tilde{r} \in \widetilde{\mathcal{R}} \}$ are one-to-one corresponded.

Assume $\mathcal{P}_1$ connects with another planar face $\mathcal{P}_2$ on a shared edge segment $\mathcal{E} = \{ \mathbf{x} \in \mathcal{Z} | \mathbf{x} \in \mathcal{P}_1, \mathbf{x} \in \mathcal{P}_2\}$. Define the normal of $\mathcal{P}_2$ as $\mathbf{n}_2 \in \mathbb{R}^3$, we have $\mathbf{n}_1 \not\propto \mathbf{n}_2$. Let $\mathcal{P}_2 \subset \mathcal{C}_F^{\tilde{r}_2}$, and we thus have $\mathcal{E} \subset \mathcal{C}_F^{\tilde{r}_1}$ and $\mathcal{E} \subset \mathcal{C}_F^{\tilde{r}_2}$, which tells that the two cells $\mathcal{C}_F^{\tilde{r}_1}$ and $\mathcal{C}_F^{\tilde{r}_2}$ connect at least on $\mathcal{E}$. Due to the monotonous and convex nature of linear analytic cells $\{ \mathcal{C}_F^{\tilde{r}} | \tilde{r} \in \widetilde{\mathcal{R}} \}$,  $\mathcal{E}$ must be on the boundaries of both $\mathcal{C}_F^{\tilde{r}_1}$ and $\mathcal{C}_F^{\tilde{r}_2}$, and the boundaries of $\mathcal{C}_F^{\tilde{r}_1}$ and $\mathcal{C}_F^{\tilde{r}_2}$ share at least on $\mathcal{E}$. There exist two cases for the connection of cell boundaries on $\mathcal{E}$: (1) in the general case, $\mathcal{C}_F^{\tilde{r}_1}$ and $\mathcal{C}_F^{\tilde{r}_2}$ share a boundary $\mathcal{B}_F^{\tilde{r}_1\tilde{r}_2}$ defined by a hyperplane $H_{lk}^{\tilde{r}_1\tilde{r}_2} = \{ \mathbf{x} \in \mathbb{R}^3 | \mathbf{a}^{\tilde{r}_1\tilde{r}_2}_{lk}\mathbf{x} = 0 \}$, and we have $\mathcal{E} \in \mathcal{B}_F^{\tilde{r}_1\tilde{r}_2}$, which, based on Corollary \ref{LemmaRegionAssociatedNeuronMapping} and Definition \ref{DefinitionNeuronMLPState}, suggests that the two cells have a switching neuron state $s_{lk}(\mathbf{x}) \ \forall \ \mathbf{x} \in \mathcal{B}_F^{\tilde{r}_1\tilde{r}_2}$, and consequently a switching neuron state $s_{lk}(\mathbf{x}) \ \forall \ \mathbf{x} \in \mathcal{E}$; (2) in some rare case, $\mathcal{E}$ coincides with a cell edge of $\mathcal{C}_F^{\tilde{r}_1}$ defined by $\{ \mathbf{x} \in \mathbb{R}^3 | \mathbf{a}^{\tilde{r}_1}_{l_1 k_1}\mathbf{x} = 0, \mathbf{a}^{\tilde{r}_1}_{l_1' k_1'}\mathbf{x} = 0 \}$,  and a cell edge of $\mathcal{C}_F^{\tilde{r}_2}$ defined by $\{ \mathbf{x} \in \mathbb{R}^3 | \mathbf{a}^{\tilde{r}_2}_{l_2 k_2}\mathbf{x} = 0, \mathbf{a}^{\tilde{r}_2}_{l_2' k_2'}\mathbf{x} = 0 \}$, and it is not necessary that $l_1k_1$ and $l_2k_2$ specify the same neuron, and $l_1'k_1'$ and $l_2'k_2'$ specify another same neuron. Due to a phenomenon similar to the blessing of (high) dimensionality \cite{DimBlessing}, the second case of coincidence is expected to happen with a low probability. In any of the two cases, the boundaries $\mathcal{C}_F^{\tilde{r}_1}$ and $\mathcal{C}_F^{\tilde{r}_2}$ respectively associated with $\mathcal{P}_1$ and $\mathcal{P}_2$ connect on $\mathcal{E}$.

Since for any pair of planar faces $\mathcal{P}_1$ and $\mathcal{P}_2$ connected on $\mathcal{Z}$, we prove that they are uniquely corresponded to a pair of analytic faces $\mathcal{P}_F^{\tilde{r}_1}$ and $\mathcal{P}_F^{\tilde{r}_2}$, which are connected via boundaries of their respective analytic cells $\mathcal{C}_F^{\tilde{r}_1}$ and $\mathcal{C}_F^{\tilde{r}_2}$. The theorem is proved.
\end{proof}

\section{More qualitative results}\label{AppendixAdditionalQualityResults}

\begin{figure*}[!h]
    \vskip 0.1in
    \begin{center}
    \includegraphics[scale=0.12]{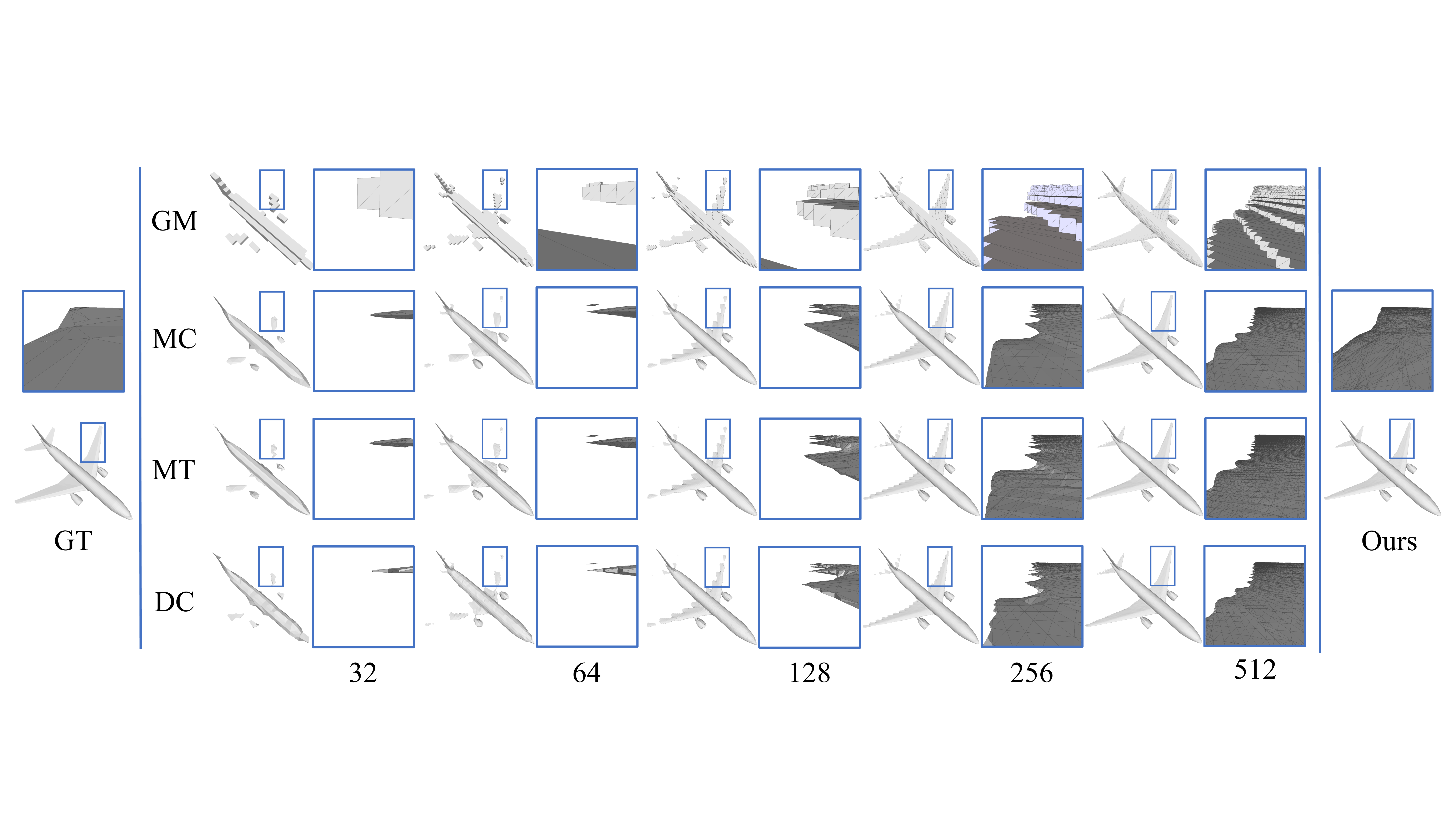}
    \caption{Qualitative comparisons of different meshing algorithms. For greedy meshing (GM), marching cubes (MC), marching tetrahedra (MT), and dual contouring (DC), results under a resolution range of discrete point sampling from $32^3$ to a GPU memory limit of $512^3$ are presented. }
    \label{FigQualityResults1}
    \end{center}
    \vskip -0.1in
\end{figure*}

\begin{figure*}[!h]
    \vskip 0.1in
    \begin{center}
    \includegraphics[scale=0.123]{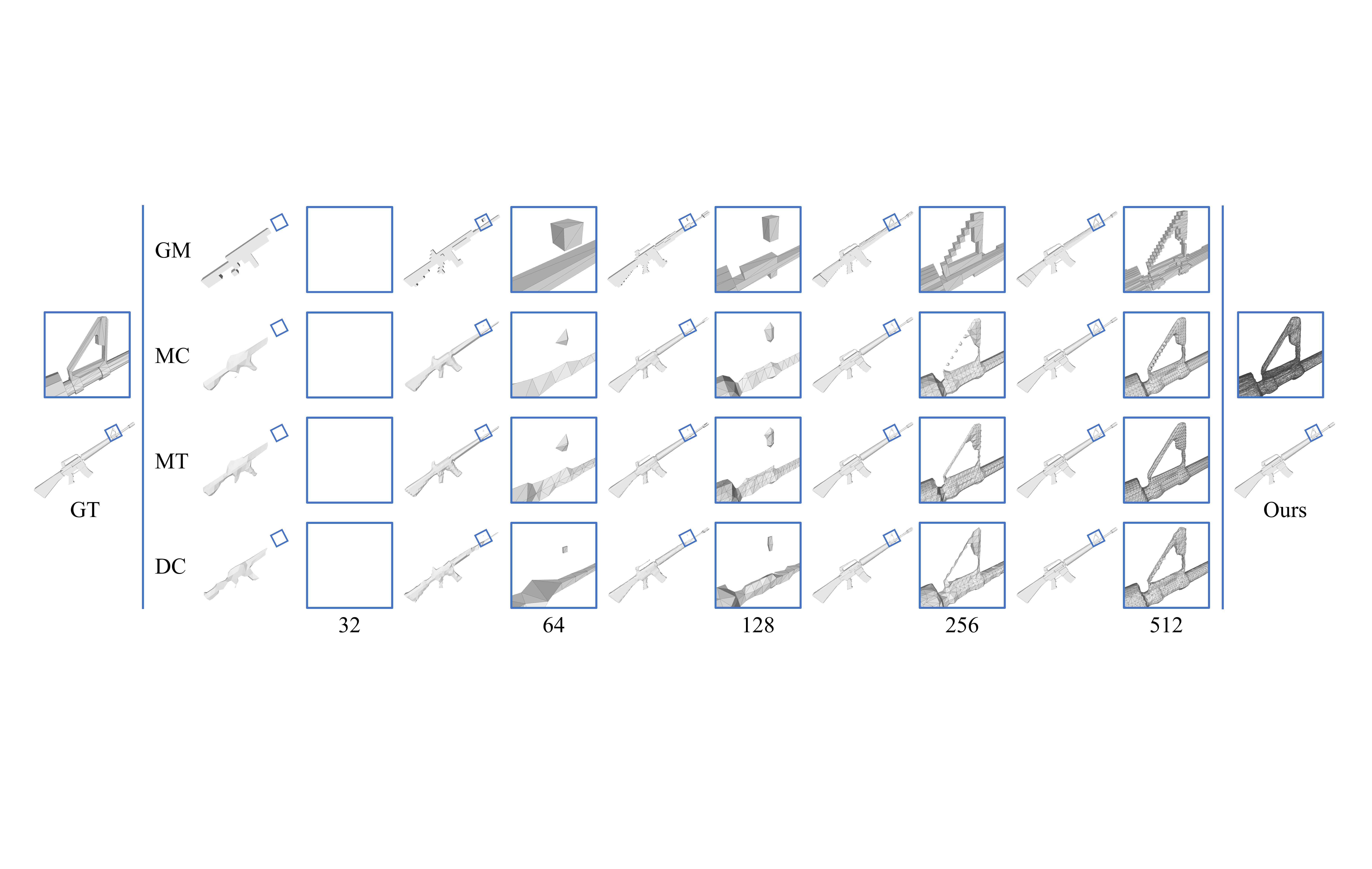}
    \caption{Qualitative comparisons of different meshing algorithms. For greedy meshing (GM), marching cubes (MC), marching tetrahedra (MT), and dual contouring (DC), results under a resolution range of discrete point sampling from $32^3$ to a GPU memory limit of $512^3$ are presented. }
    \label{FigQualityResults2}
    \end{center}
    \vskip -0.1in
\end{figure*}

\begin{figure*}[!h]
    \vskip 0.1in
    \begin{center}
    \includegraphics[scale=0.135]{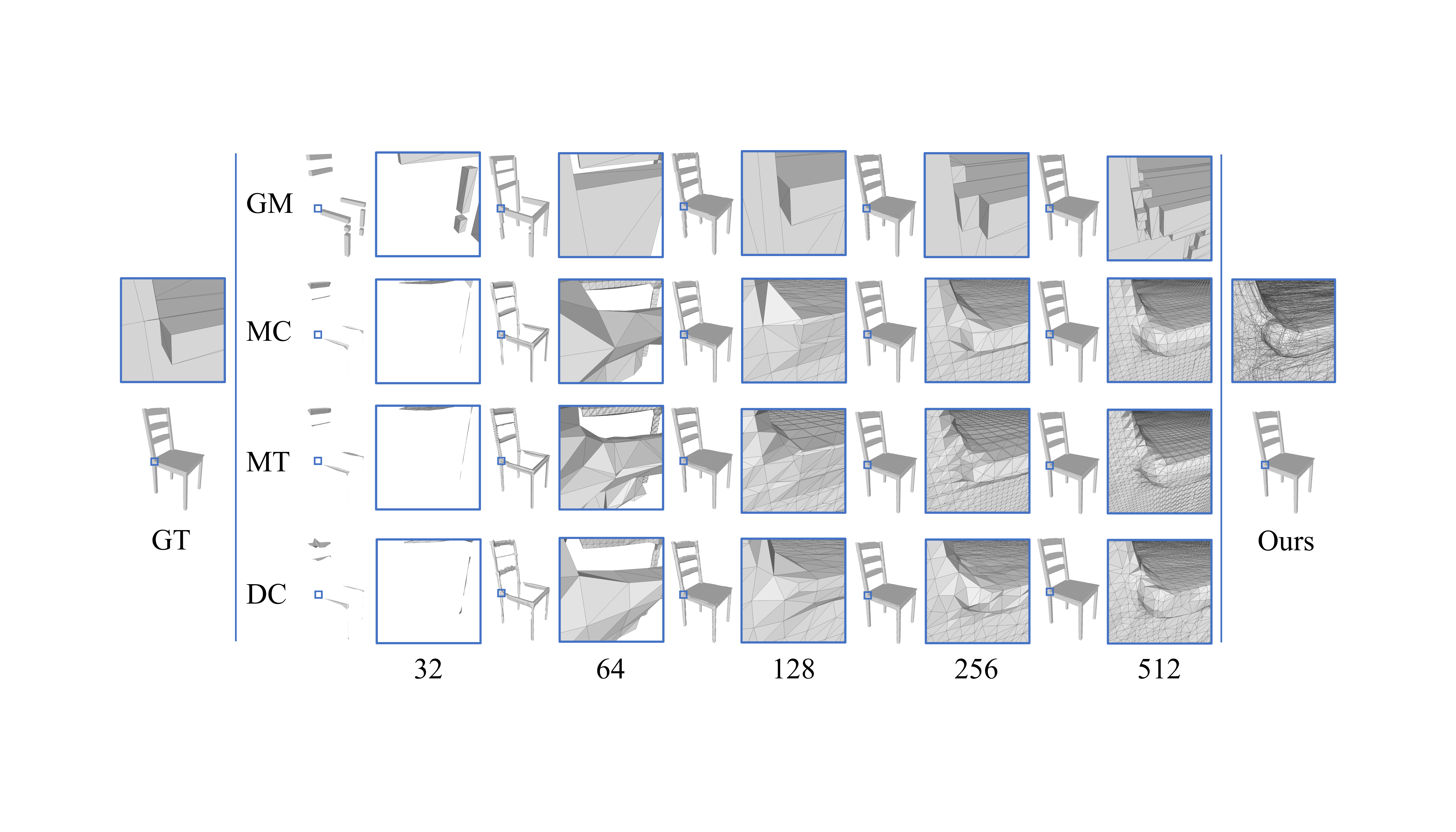}
    \caption{Qualitative comparisons of different meshing algorithms. For greedy meshing (GM), marching cubes (MC), marching tetrahedra (MT), and dual contouring (DC), results under a resolution range of discrete point sampling from $32^3$ to a GPU memory limit of $512^3$ are presented. }
    \label{FigQualityResults3}
    \end{center}
    \vskip -0.1in
\end{figure*}

\begin{figure*}[!h]
    \vskip 0.1in
    \begin{center}
    \includegraphics[scale=0.13]{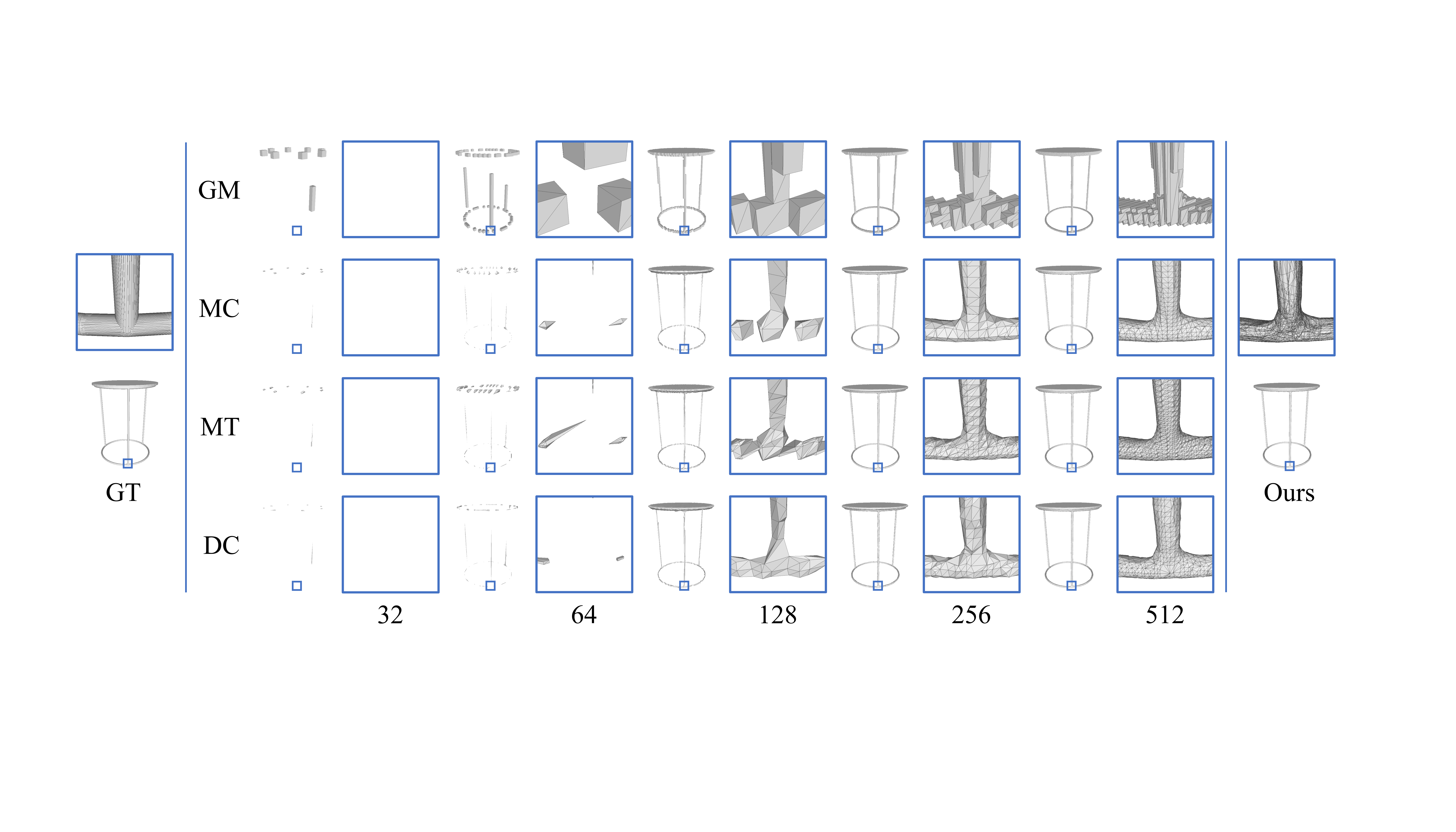}
    \caption{Qualitative comparisons of different meshing algorithms. For greedy meshing (GM), marching cubes (MC), marching tetrahedra (MT), and dual contouring (DC), results under a resolution range of discrete point sampling from $32^3$ to a GPU memory limit of $512^3$ are presented. }
    \label{FigQualityResults4}
    \end{center}
    \vskip -0.1in
\end{figure*}

We show additional qualitative results for the categories of ``Rifle'', ``Chair'', ``Airplane'', and ``Table'' in Figures \ref{FigQualityResults1}, \ref{FigQualityResults2}, \ref{FigQualityResults3}, and \ref{FigQualityResults4}.

\section{Numerical results}\label{AppendixNumericalComp}

We show in Table \ref{TableAlgorithms} the numerical results corresponding to the plotted curves in Figure \ref{FigQuantitativeCurves}.

\begin{table*}[!htbp]
    \caption{Numerical results of different meshing algorithms under metrics of recovery precision and inference world-clock time.
    For greedy meshing (GM), marching cubes (MC), marching tetrahedra (MT), and dual contouring (DC), results under a resolution range
    of discrete point sampling from $32^3$ to a GPU memory limit of $512^3$ are presented, and the dominating computations of their sampled
    points' SDF values are implemented on GPU.}
    \label{TableAlgorithms}
    \vskip 0.1in
    \begin{center}
    \begin{small}
    \begin{tabular}{cccccccc}
        \toprule
            Algorithms & CD($\times10^{-1}$) & EMD($\times10^{-3}$) & IoU($\%$) & F@$\tau$($\%$) & F@$2\tau$($\%$) & Time(sec.) \\
        \midrule
            MC32  & 37.280 & 25.368 & 72.266 & 39.805 & 78.332 & 2.25 \\
            MC64  & 6.4457 & 8.8759 & 88.461 & 61.098 & 96.116 & 2.41 \\
            MC128 & 5.5740 & 6.7293 & 91.348 & 66.585 & 97.194 & 3.41 \\
            MC256 & 5.5731 & 6.5415 & 91.410 & 66.773 & 97.202 & 14.0 \\
            MC512 & 5.5730 & 6.5403 & 91.445 & 66.777 & 97.205 & 156 \\
        \midrule
            GM32  & 45.599 & 23.918 & 63.356 & 19.725 & 55.441 & 2.41 \\
            GM64  & 11.006 & 11.259 & 76.122 & 32.580 & 88.112 & 2.50 \\
            GM128 & 6.8142 & 8.0922 & 85.078 & 53.570 & 96.793 & 3.40 \\
            GM256 & 5.9424 & 6.9674 & 88.446 & 63.411 & 97.008 & 14.2 \\
            GM512 & 5.7248 & 6.7480 & 90.098 & 65.548 & 97.015 & 171 \\
        \midrule
            MT32  & 38.485 & 25.383 & 73.677 & 41.673 & 79.961 & 2.62 \\
            MT64  & 6.5388 & 8.8540 & 88.628 & 61.548 & 96.088 & 3.47 \\
            MT128 & 5.6575 & 6.6818 & 91.306 & 66.691 & 97.228 & 7.33 \\
            MT256 & 5.5276 & 6.6185 & 91.335 & 66.994 & 97.236 & 29.5 \\
            MT512 & 5.5109 & 6.6117 & 91.347 & 66.995 & 97.239 & 204 \\
        \midrule
            DC32  & 41.570 & 28.735 & 70.134 & 37.162 & 74.564 & 2.46 \\
            DC64  & 6.9833 & 9.9407 & 87.627 & 59.704 & 95.506 & 2.61 \\
            DC128 & 5.6615 & 6.7204 & 91.304 & 66.562 & 97.215 & 3.76 \\
            DC256 & 5.5449 & 6.6735 & 91.349 & 66.921 & 97.220 & 16.2 \\
            DC512 & 5.5421 & 6.6165 & 91.355 & 66.927 & 97.221 & 177 \\
        \midrule
            Ours  & 5.5049 & 6.5401 & 91.451 & 67.153 & 97.239 & 20.8 \\
        \bottomrule
    \end{tabular}
    \end{small}
    \end{center}
    \vskip -0.1in
\end{table*}

\end{document}